\documentclass[runningheads, envcountsame, a4paper]{llncs}
\usepackage{graphicx}
\usepackage{subcaption}
\usepackage{multirow}
\usepackage[misc]{ifsym}
\usepackage[hyphens]{url}

\usepackage{cite}

\newcommand\Tstrut{\rule{0pt}{2.6ex}}  

\begin{document}
%
% Alternatively: Topological Radar and Radar-to-LiDAR Localization Using Rotational Invariant Descriptors
\title{Large-Scale Topological Radar Localization Using Learned Descriptors
\thanks{The project was funded by POB Research Centre for Artificial Intelligence and Robotics of Warsaw University of Technology within the Excellence Initiative Program - Research University (ID-UB)}}
\titlerunning{Topological Radar Localization}

\toctitle{Large-Scale Topological Radar Localization Using Learned Descriptors}

\author{Jacek Komorowski\inst{1}\orcidID{0000-0001-6906-4318} (\Letter) \and
Monika Wysoczanska\inst{1}\orcidID{0000-0001-7785-2277} \and
Tomasz Trzcinski \inst{1}\orcidID{0000-0002-1486-8906}}
\authorrunning{J. Komorowski et al.}

\tocauthor{Jacek Komorowski, Monika Wysoczanska, Tomasz Trzcinski}

\institute{Warsaw University of Technology, Warsaw, Poland
\email{\{firstname,lastname\}@pw.edu.pl}
}

\maketitle              % typeset the header of the contribution

\setcounter{footnote}{0} 

\begin{abstract}
In this work, we propose a method for large-scale topological localization based on radar scan images using learned descriptors. 
We present a simple yet efficient deep network architecture to compute a rotationally invariant discriminative global descriptor from a radar scan image.
The performance and generalization ability of the proposed method is experimentally evaluated on two large scale driving datasets: MulRan and Oxford Radar RobotCar.
Additionally, we present a comparative evaluation of radar-based and LiDAR-based localization using learned global descriptors.
Our code and trained models are publicly available on the project website.
\footnote{\url{https://github.com/jac99/RadarLoc}}
\end{abstract}

\section{Introduction}

%Place recognition is an important problem in robotics and autonomous driving community since it's at the core of Simultaneous Localization and Mapping (SLAM) systems. It aims at recognizing already visited places, which in a mapping scenario is being cast as a loop-closure detection problem. In the localization scenario, place recognition can be seen as global localization problem in e.g. limited GPS-based positioning environments~\cite{10.3389/frobt.2021.661199}.

Place recognition is an important problem in robotics and autonomous driving community. It aims at recognizing previously visited places based on an input from a sensor, such an RGB camera or a LiDAR scanner, installed on the moving vehicle.
Place recognition plays an important part in mapping and localization methods. 
It allows detecting and closing loops during a map creation. It can improve localization accuracy in areas with restricted or limited GPS coverage~\cite{10.3389/frobt.2021.661199}.

Sensing capabilities of modern robotic and autonomous driving solutions constantly improve as  technology advances and becomes more affordable~\cite{s21062140}. 
Therefore, methods for visual place recognition span from classical approaches using RGB camera images to 360$^\circ$ range-based measuring systems, such as LiDARs and radars. 
Appearance-based methods~\cite{arandjelovic2016netvlad, doi:10.1177/0278364908090961} leverage fine details of observed scenes, such as a texture of visible buildings. However, they fail under light-condition variance and seasonal changes.
Structure information-based place recognition methods address these limitations.  
Modern 3D LiDARs, such as Velodyne HDL-64E, can capture up to 100 meters range providing a rich geometrical information about the observed scene. 
LiDAR-based topological localization using learned descriptors is currently an active field of research, with a larger number of published methods~\cite{angelina2018pointnetvlad, kim2018scan, zhang2019pcan, liu2019lpd, Komorowski_2021_WACV, Komorowski_2021_MinkLocMultimodal, du2020dh3d, xu2021disco,xia2020soe}.
Nevertheless, high-end LiDARs are too expensive to be widely applicable, with the price as high as 70k USD per unit. 
Additionally, LiDAR readings are adversely affected by extreme environmental conditions such as fog, heavy rain or snow~\cite{gskim-2020-icra}. 
In such challenging conditions, radars show a great potential as they are more robust against atmospheric phenomena. 
Moreover, modern frequency-modulated continuous wave (FMCW) scanning radars cover broader area, having an operating range up to 200 meters.
However, radar-based localization is relatively little exploited~\cite{gskim-2020-icra,suaftescu2020kidnapped}.
This can be attributed to limited availability of sufficiently large and diverse datasets.
The situation has improved recently with the release of large-scale MulRan~\cite{gskim-2020-icra} and Radar RobotCar~\cite{RadarRobotCarDatasetICRA2020} datasets.

%Since each of sensor technologies mentioned above has its limitations, intuitively in order to build robust autonomous driving systems, multi-sensory method is a feasible approach, e.g. building a map in stable conditions using one modality, and then localizing in more demanding cases with complementary modality. 

\begin{figure}[tb]
\centering
\includegraphics[width=1.0\linewidth,trim={1cm 7.4cm 6.5cm 1.2cm},clip]{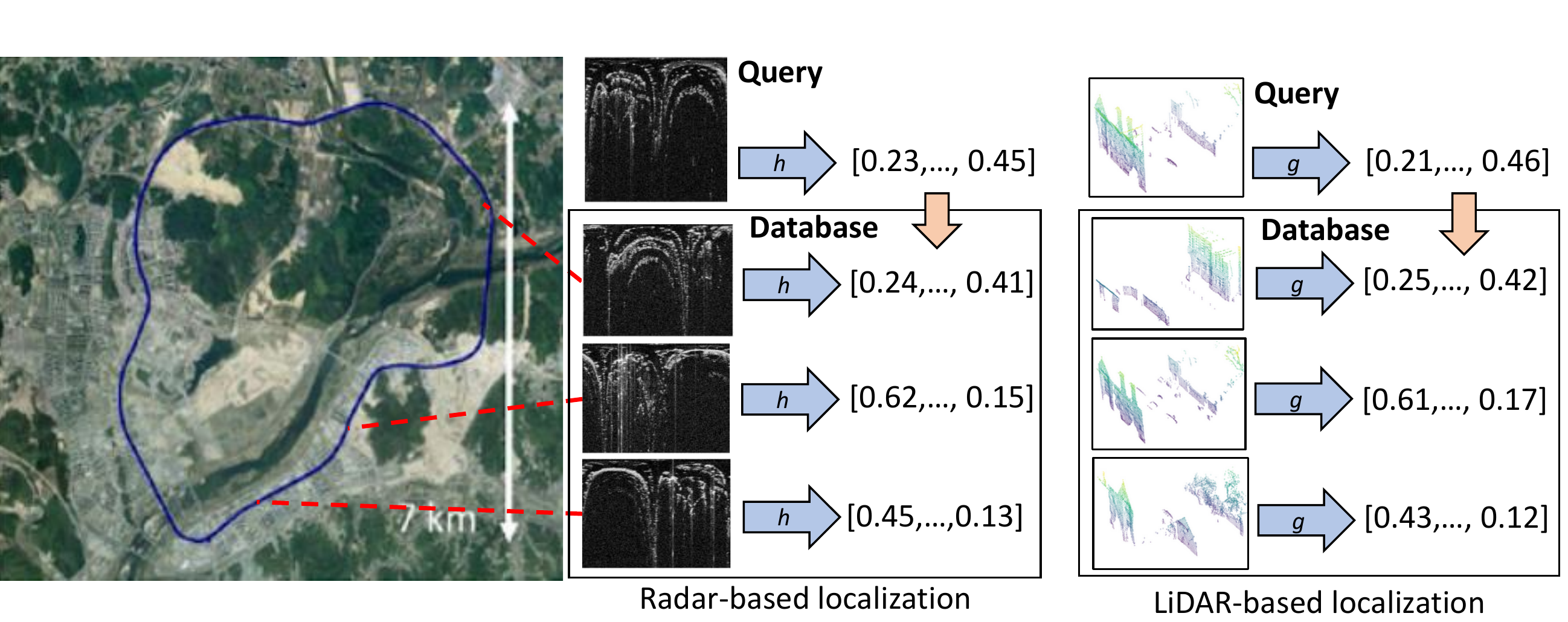}
\caption{Radar-based and LiDAR-based topological localization. Trained neural network (blue arrow) is used to compute a global descriptor from a query reading. Localization is performed by searching the database for a geo-tagged sensor readings with closest descriptors.}
\label{fig:teaser}
\end{figure}

In this work, we propose a method for topological localization using learned descriptors based on Frequency-Modulated Continuous Wave (FMCW) radar scan images. The idea is illustrated in Fig.~\ref{fig:teaser}.
A trained neural network computes low-dimensional descriptors from sensor readings.
Localization is performed by searching the database for geotagged elements with descriptors closests, in Euclidean distance sense, to the descriptor of the query scan. 
We present a simple and efficient deep network architecture to compute rotationally invariant discriminative global descriptor from a radar scan image. Rotational invariance is an important property for place recognition tasks, as the same place can be revisited from different directions. 
%Our method gives state of the art results on two large scale driving datasets: MulRan~\cite{gskim-2020-icra} and Oxford Radar RobotCar~\cite{RadarRobotCarDatasetICRA2020}. 

An interesting research question is the comparison of performance of radar-based and LiDAR-based localization methods.
%The comparison of radar-based and LiDAR-based topological localization using learned descriptors is an interesting research question.
In this work we perform such analysis by comparing our radar-based descriptor with
%state of the art 
LiDAR-based global descriptor. 
%The comparison is done using MuRan dataset, containing data gathered during multiple traversals by a vehicle equipped with both LiDAR and FMCW radar sensors. 
Contributions of this work can be summarized as follows. First, we propose a simple yet efficient deep neural network architecture to compute a discriminative global descriptor from a radar scan image. 
Second, present a comparative evaluation of radar-based and LiDAR-based localization using learned descriptors.

%We evaluate the performance of topological localization based on scan images from FMCW radar versus 3D point clouds from 360$^\circ$ 3D LiDAR on two large and diverse datasets.

\section{Related work}

\textbf{Radar-based place recognition.}
Radar-based place recognition using learned descriptors is relatively unexplored area.
In~\cite{gskim-2020-icra} authors adapt a hand-crafted ScanContext~\cite{kim2018scan} descriptor, originally used for point cloud-based place recognition.
~\cite{suaftescu2020kidnapped} presents a learned a global descriptor computed using convolutional neural network with NetVLAD~\cite{arandjelovic2016netvlad} aggregation layer, commonly used in visual domain. In order to achieve rotational invariance authors introduce couple of modifications, such as cylindrical convolutions, anti-aliasing blurring, and azimuth-wise max-pooling. 
\cite{s20216002} extends this method by developing a two-stage approach which integrates global descriptor learning with precise pose estimation using spectral landmark-based techniques. 

%In~\cite{10.3389/frobt.2021.661199} authors propose a heterogeneous approach and design a deep neural network to extract the feature embeddings of radar and LiDAR. The input to the network is ScanContext-based representation for both radar and LiDAR data. Finally, the network is trained with triplet loss. 
% TODO: compare their r2l scenario with ours

%\todo[inline]{Papers on radar-lidar localization that could be included: Radar-on-Lidar: metric radar localization on prior lidar maps; RaLL: End-to-end Radar Localization on Lidar Map Using Differentiable Measurement Model; Radar Localization and Mapping for Indoor Disaster Environments via Multi-modal Registration to Prior LiDAR Map}

\textbf{LiDAR-based place recognition.}
Place recognition methods based on LiDAR scans can be split into two categories: handcrafted and learned descriptors. Among the first group, one of the most effective methods is ScanContext~\cite{kim2018scan}, which represents a 3D point cloud as a 2D image.
3D points above the ground plane level are converted to egocentric polar coordinates and projected to a 2D plane. This idea was extended in a couple of later works~\cite{fan2020seed, cai2021weighted}.

The second group of methods leverages deep neural networks in order to compute a discriminative global descriptor in a learned manner. 
One of the first is PointNetVLAD~\cite{angelina2018pointnetvlad}, which uses PointNet architecture to extract local features, followed by NetVLAD pooling producing a global descriptor.
It was followed by a number of later works~\cite{zhang2019pcan, liu2019lpd,xia2020soe,Komorowski_2021_WACV} using the same principle: local features extracted from the 3D point cloud are pooled to yield a discriminative global descriptor.
However all these works operate on relatively small point clouds constructed by accumulating and downsampling a few consecutive scans from a 2D LiDAR.
Thus, they do not scale well to larger point clouds generated by a single 360$^\circ$ sweep from a 3D LiDAR with an order of magnitude more points.
To mitigate this limitation another line of methods uses an intermediary representation of an input point cloud before feeding it to a neural network. DiSCO~\cite{xu2021disco} first converts a point cloud into a multi-layered representation, then uses a convolutional neural network to extract features in a polar domain and produce a global descriptor.

\section{Topological localization using learned descriptors}

The idea behind a topological localization using learned descriptors is illustrated in Fig.~\ref{fig:teaser}.
A trained neural network is used to compute a discriminative global descriptor from a sensor reading (radar scan image or 3D point cloud from LiDAR). Place recognition is performed by searching the database of geo-tagged sensor readings for descriptors closests, in Euclidean distance sense, to the descriptor of the query reading. 
The geo-position of the reading with a closest descriptor found in the database approximates query location.
This coarse localization may be followed by a re-ranking step and a precise 6DoF pose estimation based on local descriptors. But in this work we focus only on coarse-level localization using learned global descriptors.

\subsection{Radar scan-based global descriptor}

\begin{figure}
\centering
\includegraphics[width=1.0\linewidth,trim={0cm 0cm 0cm 0cm},clip]{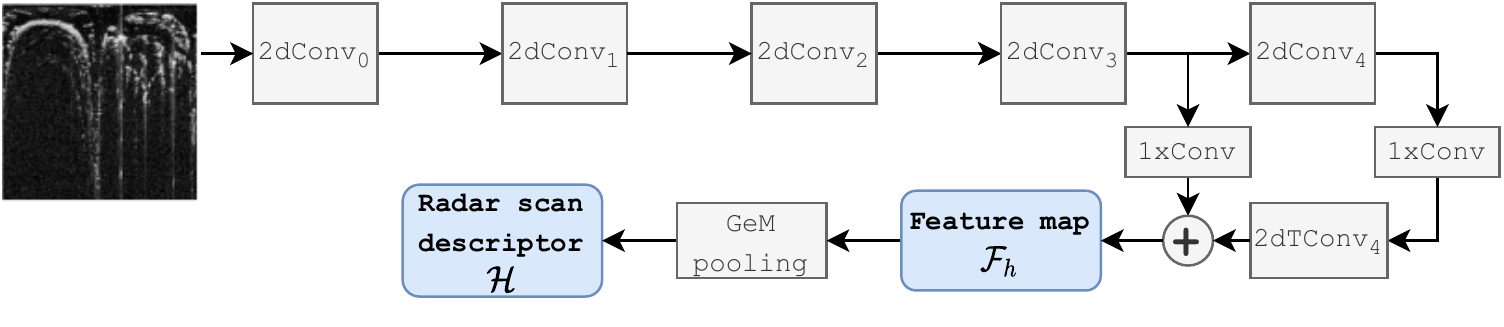}
\caption{Architecture of a radar scan descriptor extraction network.
The input is processed by a 2D convolutional network with FPN~\cite{lin2017feature} architecture to produce a feature map 
$\mathcal{F}_h$.
The feature map is pooled with generalized-mean (GeM)~\cite{radenovic2018fine} pooling to produce a radar scan descriptor $\mathcal{H}$.}
\label{fig:radar_descriptor}
\end{figure}

This section describes the architecture of our network to compute a discriminative global descriptor of an input radar scan image.
The high-level network architecture is shown in Fig.~\ref{fig:radar_descriptor}. 
It consists of two parts: local feature extraction network followed by the generalized-mean (GeM)~\cite{radenovic2018fine} pooling layer. Our initial experiments proved that GeM yields  better results compared to commonly used NetVLAD~\cite{arandjelovic2016netvlad} pooling. One of the reasons is much smaller number of learnable parameters that reduces the risk of overfitting to a moderately-sized training set.
The input to the network is a single-channel radar scan image in polar coordinates. 
The scan image is processed using a 2D convolutional network modelled after FPN~\cite{lin2017feature} design pattern.
Upper part of the network, with left-to-right data flow, contains five convolutional blocks producing 2D feature maps with
decreasing spatial resolution and increasing receptive field.
The bottom part, with right-to-left data flow, contains a transposed convolution generating an upsampled feature map. 
Upsampled feature map is concatenated with the skipped features from the corresponding block in the upper pass using a lateral connection. 
Such design is intended to produce a feature map with higher spatial resolution, having a large receptive field.
Our experiments proved its advantage over a simple convolutional architecture with one-directional data flow.
Table~\ref{jk:radar-details} shows details of each network block. The first convolutional block ({\tt 2dConv$_0$}) has a bigger 5x5 kernel, in order to aggregate information from a larger neighbourhood. Subsequent blocks ({\tt 2dConv$_1 \ldots$2dConv$_4$})
are made of a stride two convolution, which decreases spatial resolution by two, followed by residual block consisting
of two convolutional layers with 3x3 kernel and ECA~\cite{Wang_2020_CVPR} channel attention layer.
All convolutional layers are followed by batch normalization~\cite{pmlr-v37-ioffe15} layer and ReLU non-linearity. 
Two {\tt 1xConv} blocks have the same structure, both contain a single convolutional layer with 1x1 kernel. The aim of these
blocks is to unify the number of channels in feature maps produced by the blocks in the left-to-right, before they are merged with feature maps from the right-to-left pass through the network. 
The bottom part of the network (left-to-right pass)consists of a single transposed convolution layer ({\tt 2dTConv$_{4}$}) with 2x2 kernel and stride 2.
The feature map $\mathcal{F}_h$ computed by the feature extraction network is pooled with generalized-mean (GeM)~\cite{radenovic2018fine} pooling to produce a radar scan descriptor $\mathcal{H}$

Important property for loop closure applications is rotational invariance, as the same place may be revisited from different directions. 
Invariance to the viewpoint rotation translates into shift invariance along the angular direction of the scan image in polar coordinates.  
To ensure this invariance we use a circular padding along the angular direction in all convolutions in the radar scan descriptor extraction network.
The left boundary of the scan image (corresponding to $0^{\circ}$ angular coordinate) is padded with values on the right boundary of the image (corresponding to  $360^{\circ}$ angle) and vice verse.
This only partially solves the problem, as convolutions and pooling with stride 2 are not translational invariant due to the aliasing effect.
To mitigate this problem we augment the training data, by randomly rolling the image over axis corresponding to the angular direction, which gives the same effect as rotating an image in Cartesian coordinates.
Alternative approach is to apply anti-aliasing blurring~\cite{zhang2019making}, where stride 2 max-pooling and convolutions are replaced with stride 1 operations, followed by a stride 2 Gaussian blur. But we found it giving worse results in practice.

\begin{table}
	\centering
    \begin{tabular}{@{\enspace}l@{\enspace}|@{\enspace}l@{\enspace}}
     \hline
    Block  &  Layers\\
     \hline
    %\multicolumn{2}{c}{\textbf{Bottom-up trunk}} \\
    {\tt 2dConv$_0$}  &  2d convolutions with 32 filters 5x5 - BN - ReLU\\
%    {\tt conv$_k$}, $k = 1, \ldots, 7$  &  32 filters 2x2x2 stride 2   \\
    \Tstrut
    {\tt 2dConv$_k$} &  2d convolution with $c_k$ filters 2x2 stride 2  - BN - ReLU \\
           &  2d convolution with $c_k$ filters 3x3 stride 1 - BN - ReLU  \\
           &  2d convolution with $c_k$ filters 3x3 stride 1 - BN - ReLU\\
           & Efficient Channel Attention (ECA) layer~\cite{Wang_2020_CVPR}  \\
           & where $c_1=32, c_{2,3} = 64, c_{4,5} = 128$ \\
    %\hline
    %\multicolumn{2}{c}{\textbf{Global branch}} \\
    {\tt 2dTConv$_6$} &  2d transposed conv. with 128 filters 2x2 stride 2  \\
    \Tstrut
    {\tt 1xConv} &  2d convolution with 128 filters 1x1 stride 1  \\
    \Tstrut
    {\tt GeM pooling} & Generalized-mean Pooling layer~\cite{radenovic2018fine} \\
    \hline
    \end{tabular}
	\caption{Details of the descriptor extractor network for radar scan images.}
    \label{jk:radar-details}
\end{table}

\subsection{LiDAR-based global descriptor}

This section describes the architecture of the network used to compute a discriminative global descriptor of a 3D point cloud. 
We choose a 3D convolutional architecture using sparse volumetric representation that produced state of the art results in our previous MinkLoc3D~\cite{Komorowski_2021_WACV} work. 
However, MinkLoc3D is architectured to process relatively small point clouds, constructed by concatenating multiple 2D LiDAR scans. %Each point cloud covered 20 meter distance and contained 4096 points.
In this work we use point clouds build from a single 360$^\circ$ scans from 3D LiDAR, covering much larger area, app. 160 meters in diameter, and containing an order of magnitude more points. 
To extract informative features from such larger point clouds we enhanced the network architecture to increase the receptive field.
The number of blocks in upper and lower part of the network is increased compared to MinkLoc3D design. Fig.~\ref{fig:radar_descriptor} shows high-level architecture and details of each network block are given in Tab.~\ref{jk:radar-details}.
For more information we refer the reader to our MinkLoc3D~\cite{Komorowski_2021_WACV} paper.
To ensure rotational invariance of the resultant global descriptor, necessary for loop closure applications, we resort to data augmentation. Point clouds are randomly rotated around the $z$-axis before they are fed to the network during the training. 

%Similarly as for radar descriptor extraction, the network consists of two parts: local feature extraction network followed by the generalized-mean (GeM)~\cite{radenovic2018fine} pooling layer. 
%The input point cloud $P=\left\{ \left( x_i, y_i, z_i \right) \right\}$ is first quantized into a single channel sparse tensor $\hat{P}=\left\{ \left( \hat{x}_i, \hat{y}_i, \hat{z}_i, 1 \right) \right\}$.
%The values of this single channel are set to one for non-empty voxels.
%The sparse tensor if fed to the local feature extraction network, which produces a sparse 3D feature map $\mathcal{F}_g=\left\{ \left( \hat{x}_j, \hat{y}_j, \hat{z}_j, f_j^{(1)}, \ldots, f_j^{(c)} \right) \right\}$, where $c$ is a feature dimensionality (256 in our experiments), $\hat{x}_j, \hat{y}_j, \hat{z}_j$ quantized coordinates and $f_j^{(1)}, \ldots, f_j^{(c)}$ features of $j$-th feature map element.
%The sparse 3D feature map $\mathcal{F}_g$ is pooled using a generalized-mean (GeM) pooling~\cite{radenovic2018fine} layer, which produces a global descriptor vector $\mathcal{G}$.

\begin{figure}
\centering
\includegraphics[width=1.0\linewidth]{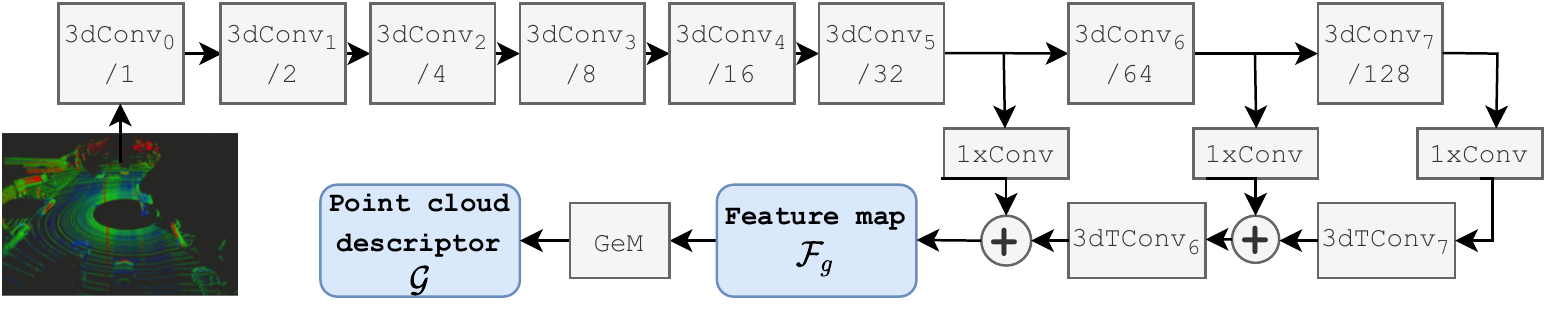}
\caption{Architecture of a LiDAR point cloud descriptor extraction network.
The input is quantized into a sparse voxelized representation and processed by a 3D convolutional network with FPN~\cite{lin2017feature} architecture. 
The resultant sparse 3D feature map is pooled with generalized-mean (GeM) pooling to produce a global point cloud descriptor $\mathcal{G}$.
} 
\label{fig:LiDAR_descriptor}
\end{figure}

\begin{table}
	\centering
    \begin{tabular}{@{\enspace}l@{\enspace}|@{\enspace}l@{\enspace}}
     \hline
    Block  &  Layers\\
     \hline
    %\multicolumn{2}{c}{\textbf{Bottom-up trunk}} \\
    {\tt 3dConv$_0$}  &  32 filters 5x5x5 - BN - ReLU\\
%    {\tt conv$_k$}, $k = 1, \ldots, 7$  &  32 filters 2x2x2 stride 2   \\
    \Tstrut
    {\tt 3dConv$_k$} &  3d convolution with $c_k$  filters 2x2x2 stride 2  - BN - ReLU \\
           &  3d convolution with $c_k$ filters 3x3x3 stride 1 - BN - ReLU  \\
           &  3d convolution with $c_k$ filters 3x3x3 stride 1 - BN - ReLU\\
           & Efficient Channel Attention (ECA) layer~\cite{Wang_2020_CVPR}  \\
           & where $c_1=32, c_2 = c_3=64, c_{4\ldots7}=128$ \\
    %\hline
    {\tt 3dTConv$_k$}, $k=6,7$  &  3d transposed conv. with 128 filters 2x2x2 stride 2  \\
    \Tstrut
    {\tt 1xConv} &  3d convolution with 128 filters 1x1x1 stride 1  \\
    \Tstrut
    {\tt GeM pooling} & Generalized-mean Pooling layer~\cite{radenovic2018fine} \\
    \hline
	\end{tabular}
    \caption{Details of the descriptor extractor network for LiDAR point clouds.}
    \label{jk:LiDAR-details}
\end{table}

\subsection{Network training}

To train both networks to generate discriminative global descriptors we use a deep metric learning approach~\cite{lu2017deep} with a triplet margin loss~\cite{hermans2017defense} defined as:
\[
L(a_i,p_i,n_i) = \max \left\{ d(a_i,p_i) - d(a_i,n_i) + m, 0 \right\} ,
\]
where 
\(d(x,y) = || x - y ||_2\) is an Euclidean distance between embeddings $x$ and $y$;  \(a_i, p_i, n_i\) are embeddings of an anchor, a positive and a negative elements in $i$-th training triplet and $m$ is a margin hyperparameter. 
The loss is formulated to make embeddings of structurally dissimilar sensor readings (representing different places) further away, in the descriptor space, than embeddings of structurally similar readings (representing the same place).
The loss is minimized using a stochastic gradient descent with Adam optimizer. 

%The network is trained using a stochastic gradient descent approach  
%Batches contain triplets consisting of an anchor, a positive and a negative element. 
%A positive element is a point cloud taken at neighbourhood location (within 20 meter distance).
%A negative element is a point cloud showing a different place than an anchor. See Fig.~\ref{fig:high_level} for visualization.
%A randomly chosen negative element would often depict a scene that is very different, both in appearance and geometry, from an anchor element. In the presence of such easy cases, the network will quickly learn how to produce sufficiently different embeddings and the training will stagnate. 
To improve effectiveness of the training process we use batch hard negative mining~\cite{hermans2017defense} strategy to construct informative triplets. Each triplet is build using the hardest negative example found within a batch. The hardest negative example is a structurally dissimilar element that has the closest embedding, computed using current network weights, to the anchor.
%The downside of using batch hard negative mining is increased risk of a training collapse, especially when the batch size is large.
%The network is faced with too difficult examples to start learning.
%To mitigate this, similar as in~\cite{Komorowski_2021_WACV}, we use dynamic batch sizing. Initially, smaller batches are constructed (32 elements). When the network starts learning, which is manifested by increasing number of triplets with zero loss, the batch size is gradually increased up to 128 elements. 
%This prevents the training process from collapse in initial epochs and allows using larger batches later to mine more difficult triplets.

To increase variability of the training data, reduce overfitting and ensure rotational invariance of global descriptors, we use on-the-fly data augmentation. 
For radar scan images, we use random erasing augmentation~\cite{zhong2017random} and random cyclical shift over $x$-axis (angular dimension). 
For points clouds, it includes random jitter and random rotation around $z$-axis. We also adapted random erasing augmentation to remove 3D points within the randomly selected frontoparallel cuboid.

\section{Experimental Results}

\subsection{Datasets and evaluation methodology}

To train and evaluate our models we use two recently published large-scale datasets: MulRan~\cite{gskim-2020-icra} and Oxford Radar RobotCar~\cite{RadarRobotCarDatasetICRA2020}.
%Both datasets are gathered by a vehicle equipped with a suite of sensors, including 360$^\circ$ 3D LiDAR and FMCW radar.
MuRan dataset is gathered using a vehicle equipped with Ouster OS1-64 3D LiDAR with 120 m. range and Navtech CIR204-H FMCW scanning radar with 200 operating range.
Radar RobotCar data is acquired using Velodyne HDL-32E LiDAR with 100 m. range and Navtech CTS350-X scanning radar with 160 m. operating range.

In both datasets each trajectory is traversed multiple times, at different times of day and year, allowing a realistic evaluation of place recognition methods.
Radar scans are provided in similar format, as $360^{\circ}$ polar images with 400 (angular) by 3360 (radial dimension) pixel resolution. To decrease computational resources requirements we downsample them to 384 (angular) by 128 (radial) resolution.
LiDAR scans are given as unordered set of points, containing between 40-60 thousand points.
To speed up the processing, we remove uninformative ground plane points with $z$-coordinate below the ground plane level.
Both datasets contain ground truth positions for each traversal.

The longest and most diverse trajectory from MulRan dataset, Sejong, is split into disjoint training and evaluation parts.
We evaluate our models using disjoint evaluation split from Sejong trajectory and two other trajectories: KAIST and Riverside, acquired at different geographic locations.
Each evaluation trajectory contains two traversals gathered at different times of a day and year. The first traversal (Sejong01, KAIST01, Riverside01) forms a query set and the second one (Sejong02, KAIST02, Riverside02) is used as a map.
To test generalization ability of our model, we test it using two traversals from Radar RobotCar dataset acquired at different days: traversal 019-01-15-13-06-37-radar-oxford-10k as a map split and 2019-01-18-14-14-42-radar-oxford-10k as a query split.

\begin{figure}
     \centering
     \begin{subfigure}[b]{0.22\textwidth}
         \centering
         \includegraphics[height=2.2cm]{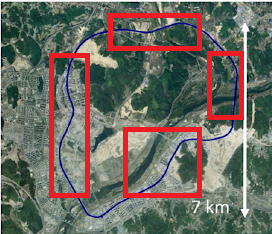}
         \caption{Sejong}
         %\label{fig:y equals x}
     \end{subfigure}
     %\hfill
     \begin{subfigure}[b]{0.29\textwidth}
         \centering
         \includegraphics[height=2.2cm]{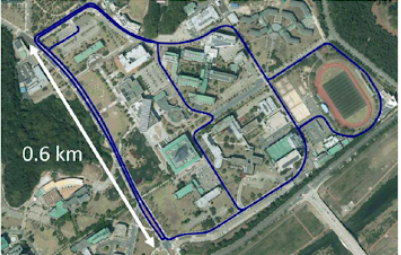}
         \caption{KAIST}
         %\label{fig:three sin x}
     \end{subfigure}
     %\hfill
     \begin{subfigure}[b]{0.29\textwidth}
         \centering
         \includegraphics[height=2.2cm]{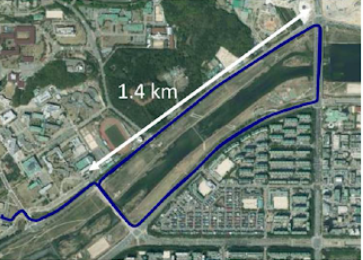}
         \caption{Riverside}
         %label{fig:five over x}
     \end{subfigure}
     %\hfill
     \begin{subfigure}[b]{0.17\textwidth}
         \centering
         \includegraphics[height=2.2cm]{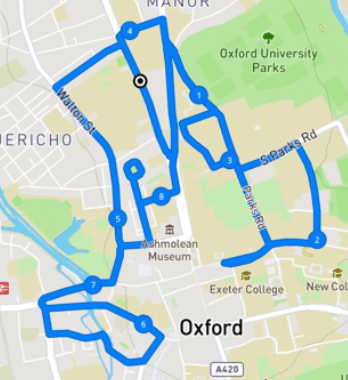}
         \caption{RobotCar}
         %label{fig:five over x}
     \end{subfigure}

        \caption{Visualization of three trajectories in MulRan dataset and one in Radar RobotCar. Red rectangles in Sejong trajectory delimit the  training split.}
        \label{fig:datasets}
\end{figure}

To avoid processing multiple scans of the same place when the car doesn't move, we ignore consecutive readings with less than 0.1m displacement in the ground truth position. We also ignore readings for which a ground truth pose in not given with 1 sec. tolerance.
See Fig.~\ref{fig:datasets} for visualization of training and evaluation trajectories.
Details of the training and evaluation sets are given in Tab.~\ref{jk:tab:splits}.

\begin{table}[tbp]
\caption{Length and number of scans in training and evaluation sets.}
\begin{center}
%@{\enspace}
\begin{tabular}{@{\enspace}l@{\enspace}|@{\enspace}c@{\enspace}r@{\enspace}|@{\enspace}c@{\enspace}|@{\enspace}c@{\enspace}}
%\hline
\multirow{2}{*}{Trajectory} & \multirow{2}{*}{Split} & \multirow{2}{*}{Length} & \multicolumn{2}{@{}c@{}}{Number of scans (map/query)} \\
 & &  & Radar & LiDAR \\
\hline
Sejong & train & 19 km & $13\,611$ & $33\,698$ \\
\hline
Sejong & test & 4 km & $1\,366$/$1\,337$ & $3\,358$/$3\,315$ \\
KAIST & test &  6 km & $3\,209$/$3\,420$ & $7\,975$/$8\,521$  \\
Riverside & test & 7 km & $2\,117$/$2\,043$ & $5\,267$/$5\,084$ \\
Radar RobotCar & test & 10 km & $5\,904$/$5\,683$ & $26\,719$/$25\,944$
\\[2pt]
\hline
\end{tabular}
\end{center}
\label{jk:tab:splits}
\end{table}

Training triplets are generated using the ground truth coordinates provided in each dataset. Positive examples is chosen from structurally similar readings, that is readings that are at most 5m apart. 
Negative examples are sampled from  dissimilar readings, that is readings that are at least 20m apart.

\textbf{Evaluation metrics.} 
To evaluate the global descriptor performance we follow similar evaluation protocol as in other point cloud-based or radar-based place recognition works~\cite{angelina2018pointnetvlad}.
Each evaluation set is split into two parts: a query and a database set, covering the same geographic area. 
A query is formed from sensor readings acquired during one traversal and the database is build from data  gathered during a different traversal, on a different day.
For each query element we find database element with closests, in Euclidean distance sense, global descriptors.
Localization is successful if at least one of the top $N$ retrieved database elements is within $d$ meters
from the query ground truth position. 
\emph{Recall@N} is defined as the percentage of correctly localized queries. 
%We report recall metric for threshold $d = 5$ and $10$ m.

%\begin{figure}
%     \centering
%     \begin{subfigure}[b]{0.45\textwidth}
%         \centering
%         %\includegraphics[height=2.6cm]{}
%         \caption{}
%         %\label{fig:three sin x}
%     \end{subfigure}
%     %\hfill
%     \begin{subfigure}[b]{0.45\textwidth}
%         \centering
%         %\includegraphics[height=2.6cm]{}
%         \caption{}
%         %label{fig:five over x}
%     \end{subfigure}
%        \caption{Exemplary radar scan from MulRan dataset. (a) Radar scan in Cartesian coordinate system (b) Radar scan in polar %coordinates system which is processed by our network. $x$-axis corresponds to the angular coordinate and $y$ to the radial %coordinate.}
%        \label{fig:dataset_items}
%\end{figure}

\subsection{Results and discussion}

\begin{figure*}[tb]
\centering
\subfloat[Sejong\label{1a}]{%
\includegraphics[height=4.2cm]{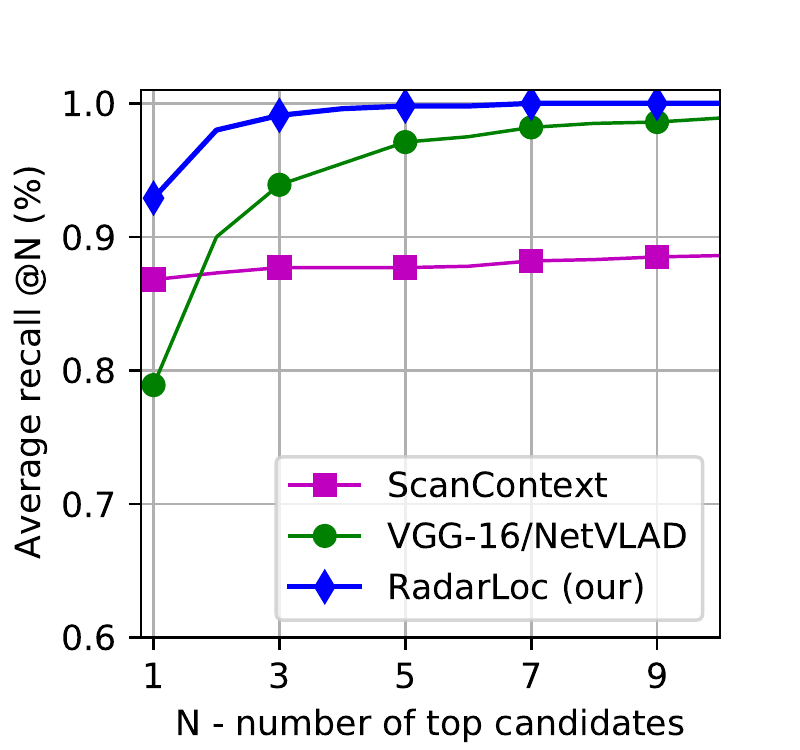}}
\hfill
\subfloat[KAIST\label{1b}]{%
\includegraphics[height=4.2cm]{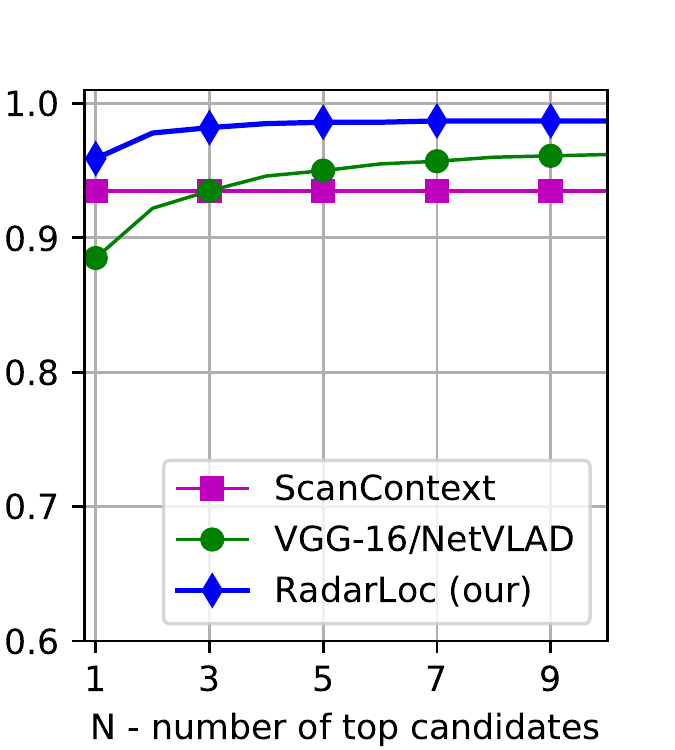}}
\hfill
\subfloat[Riverside\label{1c}]{%
\includegraphics[height=4.2cm]{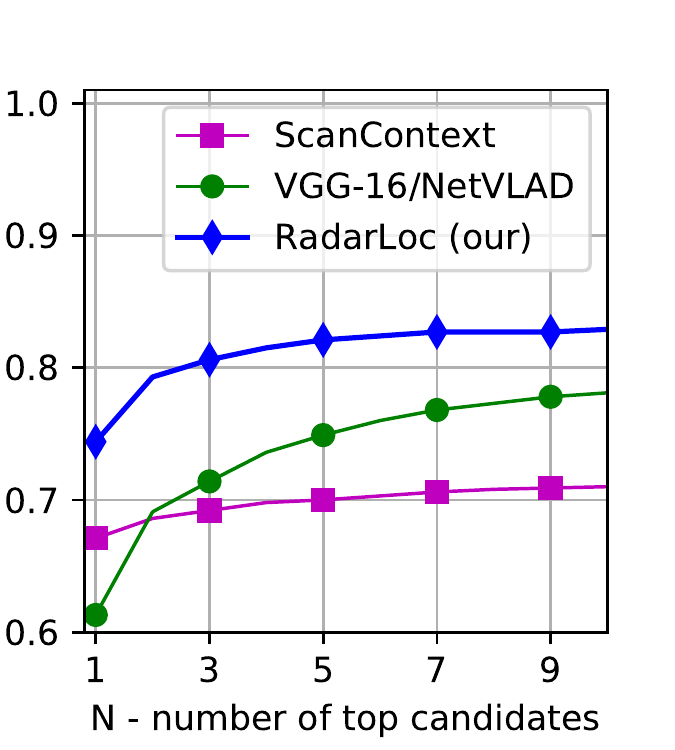}}
\caption{Average Recall@N with 5m. threshold of radar-based descriptor.}
\label{fig:plots_refined}
\end{figure*}

We compare the performance of our radar-based global descriptor with a hand-crafted ScanContext~\cite{kim2018scan} method and our re-implementation of the VGG-16/ NetVLAD architecture used as a baseline in~\cite{suaftescu2020kidnapped}.
Evaluation results are shown in Tab.~\ref{jk:tab:results1}.
Our method (RadarLoc) consistently outperforms other approaches on all evaluation sets at both 5m and 10m threshold. 
Recall@1 with 5m threshold is between 2-14 p.p. higher than the runner-up ScanContext.
At larger 10m threshold, learning-based VGG-16/NetVLAD architecture scores higher than ScanContext, but it's still about 5-10 p.p. lower than RadarLoc.
Our method generalizes well to a different dataset. The model trained on a subset of Sejong traversal, from MulRan dataset, has a top performance when evaluated on Radar RobotCar.
It must be noted that these two  datasets are acquired using a different suite of sensors, although having similar operational characteristics.
Figure~\ref{fig:plots_refined} shows Recall@$N$ plots, for $N$ ranging from 1 to 10, of all evaluated methods on different evaluation sets from MulRan dataset.
For Sejong and KAIST trajectory, the performance of our method increases with $N$, quickly reaching near-100\% accuracy.
However characteristics of the environment in which data was gathered impacts all evaluated methods.
The results for all traversals acquired in the city centre areas (Sejong, KAIST, Radar RobotCar) are relatively better, for all evaluated methods.
Riverside traversal is acquired outside the city, where significantly fewer structural elements, such as buildings or lamp posts, are present. Hence it gives considerably worse results for all evaluated methods.

\begin{table}[htbp]
\caption{Evaluation results (Recall@1) of a radar-based descriptor.}
\begin{center}
%@{\enspace}
\begin{tabular}{@{\enspace}l@{\enspace}|@{\enspace}c@{\enspace}c@{\enspace}|@{\enspace}c@{\enspace}c@{\enspace}|@{\enspace}c@{\enspace}c@{\enspace}|@{\enspace}c@{\enspace}c@{\enspace}}
%\hline
\multirow{2}{*}{Method}
&  \multicolumn{2}{@{}c@{}}{Sejong} & \multicolumn{2}{@{}c@{}}{KAIST} & \multicolumn{2}{@{}c@{}}{Riverside} & \multicolumn{2}{@{}c@{}}{RadarRobotCar} \\
&  5m. & 10m.
&  5m. & 10m.
&  5m. & 10m.
&  5m. & 10m.
\\[2pt]
\hline
%\Tstrut
% ScanContext results are for 384_128 resolution. 
%ScanContext~\cite{kim2018scan} (no rerank.) & 0.556 & 0.634 & & & &  \\ 
%ScanContext~\cite{kim2018scan} (reranking) 0.857 & 0.853 & & & & \\ 
% ScanContext results are for 120_40 resolution. 
Ring key~\cite{kim2018scan} & 0.503 & 0.594 & 0.805 & 0.838 & 0.497 & 0.595 & 0.747 & 0.786 \\ 
ScanContext~\cite{kim2018scan} & 0.868 & 0.879 & 0.935 & 0.946 & 0.671 & 0.772 & 0.906 & 0.933 \\ 
%Model radar_base1.txt weights: ../weights/model_minkg1_baseline1_20210624_1138_final.pth trained with random rotation augmentation
% model_minkg1_baseline1_20210624_2156_final.pth
VGG-16/NetVLAD & 0.789 & 0.938 & 0.8885 & 0.937 & 0.613 & 0.834 & 0.883 & 0.939 \\
%KidnappedRadar~\cite{suaftescu2020kidnapped} & &  & & & &  \\
%DiSCO~\cite{xu2021disco} & &  & & & &  \\
%model_radarnet10_4_1_20210714_1034_final.pth
RadarLoc (ours) & \textbf{0.929} & \textbf{0.988} & \textbf{0.959} & \textbf{0.988} & \textbf{0.744} & \textbf{0.923} 
& \textbf{0.949} & \textbf{0.981} 
\\[2pt]
\hline
\end{tabular}
\end{center}
\label{jk:tab:results1}
\end{table}

Table~\ref{jk:tab:radar-lidar} compares performance of radar-based and LiDAR-based localization on three different evaluation sets from MulRan dataset.
LiDAR has 2-3 times higher scanning frequency than radar and our datasets contain 2-3 more LiDAR 3D point clouds than radar scan images. 
To cater for this difference, we compare the performance of radar versus LiDAR-based topological localization in two scenarios.
First, we use all available data gathered during vehicle traversals. In this scenario, LiDAR scans cover the mapped area denser than radar.
Second, we evaluate the performance, using a smaller, subsampled LiDAR dataset with the same number of point clouds as radar scans.
In this scenario, LiDAR and radar scans cover the mapped area equally dense.
It can be seen, that better map coverage caused by higher LiDAR scanning frequency has little impact. Performance of the LiDAR-based method on full evaluation sets and on downsampled evaluation sets is very close.

Intuitively, LiDAR provides richer 3D information about the structure of the observed scene, whereas radar captures only 2D range data. We would expect LiDAR-based approach to produce more discriminative global descriptors.
Interestingly, this not happen and localization using radar scan images generally yields better results, especially with larger 10m threshold.
We hypothesize that the reason is bigger complexity and larger susceptibility to overfitting of a LiDAR-based model. 
The performance of both methods on MulRan evaluation split, having scenes with similar characteristic as in the training split, is very close.
When evaluated on different traversals, covering locations with different characteristic (e.g. suburbs versus city center), LiDAR-based model generalizes worse.

\begin{table}[htbp]
\caption{Comparison (Recall@1) of radar-based and LiDAR-based descriptors.}
\begin{center}
%@{\enspace}
\begin{tabular}{@{\enspace}l@{\enspace}|@{\enspace}c@{\enspace}c@{\enspace}|@{\enspace}c@{\enspace}c@{\enspace}|@{\enspace}c@{\enspace}c@{\enspace}}
%\hline
\multirow{2}{*}{Method}
&  \multicolumn{2}{@{}c@{}}{Sejong} & \multicolumn{2}{@{}c@{}}{KAIST} & \multicolumn{2}{@{}c@{}}{Riverside} \\
&  5m. & 10m.
&  5m. & 10m.
&  5m. & 10m.
\\[2pt]
\hline
%\Tstrut
RadarLoc (our) & 0.929 & \textbf{0.988} & \textbf{0.959} & \textbf{0.988} & 0.744 &  \textbf{0.923} 
\\
% test_L2L_Riverside01_Riverside02_384_128.pickle
% model_minkg1_radarnet10_7_1_20210628_1737_final.pth
LiDAR-based & \textbf{0.950} & 0.986 & 0.901 & 0.930 & \textbf{0.748} & 0.881
\\
LiDAR-based (subsampled dataset) &  0.941 & 0.986 & 0.897 & 0.929 & 0.740 & 0.881
\\[2pt]
\hline
\end{tabular}
\end{center}
\label{jk:tab:radar-lidar}
\end{table}

%\subsubsection{Robustness tests.} xxxx

%\begin{table}[htbp]
%\caption{Ablation study. Recall@1 with 5 and 10m. threshold averaged over three MulRan test splits.}
%\begin{center}
%@{\enspace}
%\begin{tabular}{@{\enspace}l@{\enspace}|@{\enspace}c@{\enspace}c@{\enspace}}
%\hline
%\multirow{2}{*}{Method}
%&  \multicolumn{2}{@{}c@{}}{Sejong} \\
%&  5m. & 10m.
%\\[2pt]
%\hline
%\Tstrut
% model_ablation_nocyl_20210714_1325
% No cylindrical convolutions - evaluation without rotation
%RadarLoc (ours) & \textbf{0.877} & \textbf{0.966} \\
%-- no cylindrical convolutions & 0.873 & 0.961 \\
% model_params_path: ../models/radarnet2_4.txt                                                                                     
% Model name: model_ablation_nofpn2_20210714_1422 
% No FPN 2 - 4 bottom up layers
%-- no FPN & 0.869 & 0.963 \\ 
%
%-- max pooling & 0.869 & 0.964  \\ 
%\\[2pt]
%\hline
%\end{tabular}
%\end{center}
%\label{jk:tab:results1}
%\end{table}

\section{Conclusion}

Topological localization using radar scan images is relatively unexplored area. In this work we demonstrate that it has a high potential and is competitive to LiDAR-based approaches.
Our proposed radar-based localization method, having a simple and efficient architecture, outperforms baseline methods and has better generalization abilities than LiDAR-based approach.
LiDAR, however, provides richer 3D information about the structure of the observed scene, whereas radar captures only 2D range data. An interesting research question is how to fully exploit structural information available in 3D LiDAR scans to increase discriminative power of resultant global descriptors and outperform radar-based methods.

%===========================================================================================================

%\bibliographystyle{splncs04}
%\bibliography{jk-bib}

\end{document}